\documentclass{article}

\usepackage{arxiv}

\usepackage[utf8]{inputenc}
\usepackage[T1]{fontenc}
\usepackage{hyperref}
\usepackage{url}
\usepackage{booktabs}
\usepackage{amsfonts}
\usepackage{amsmath}
\usepackage{amssymb}
\usepackage{nicefrac}
\usepackage{microtype}
\usepackage{graphicx}
\usepackage{natbib}
\usepackage{doi}
\usepackage{multirow}
\usepackage{array}
\usepackage{longtable}
\usepackage{xcolor}
\usepackage{subcaption}
\usepackage{float}
\usepackage{listings}
\usepackage{inconsolata}

\newcommand{\benchmark}{UserToolBench}
\newcommand{\pcal}{\mathcal{P}}
\newcommand{\tcal}{\mathcal{T}}
\newcommand{\hcal}{\mathcal{H}}

\title{\benchmark: A User-Profile-Hidden Benchmark for Personalized Decision Making in Tool-Use LLMs}

\author{
Xuexiong Yin\textsuperscript{1} \quad
Zechuan Chen\textsuperscript{1} \quad
Yongsen Zheng\textsuperscript{2} \quad
\\[0.35em]
Yuxiang Zhang\textsuperscript{3} \quad
Jingyuan Yang\textsuperscript{3} \quad
Bin Wang\textsuperscript{3} \quad
Yubin Wang\textsuperscript{3,\textdagger} \quad
Keze Wang\textsuperscript{1,\textdagger}
\\[0.7em]
\textsuperscript{1}Sun Yat-Sen University
\qquad
\textsuperscript{2}Nanyang Technological University
\\[0.25em]
\textsuperscript{3}Huawei Noah's Ark Lab
\\[0.45em]
\textsuperscript{\textdagger}Corresponding authors
}

\date{}

\renewcommand{\shorttitle}{UserToolBench}

\hypersetup{
  hidelinks,
  pdftitle={UserToolBench: A User-Profile-Hidden Benchmark for Personalized Decision Making in Tool-Use LLMs},
  pdfauthor={
Xuexiong Yin, Zechuan Chen, Yongsen Zheng, Yuxiang Zhang,
Jingyuan Yang, Bin Wang, Yubin Wang, Keze Wang
},
  pdfsubject={Under review at EMNLP 2026},
  pdfkeywords={personalization, tool-use language models, benchmark, user profiles, decision making}
}

\begin{document}
\maketitle

\begin{abstract}
Tool-use LLMs are increasingly asked to act on users' behalf, but existing benchmarks usually focus on profile recall, style imitation, generic tool use, or response-level personalization. We introduce \benchmark\ , a benchmark for personalized decision making in tool-use LLMs. \benchmark\ tests whether a model can infer latent user preferences from interaction history, recognize when clarification is needed, and produce user-aligned tool-call trajectories under incomplete information. The benchmark is built from privacy-sanitized real interaction traces and combines structured persona profiles, public API-style tool ecosystems, and long-horizon multi-turn trajectories. It includes 10 user profiles, 36 tool sets, 1,065 turns, 170 unique tools, and evaluation-focused task types covering lack-of-information, single-tool, and multi-tool settings. Experiments with strong tool-use LLMs show that current models still have difficulty with personalized delegation. Multi-tool coordination, missing-constraint inference, and long-horizon behavioral consistency remain major bottlenecks. These results suggest that personalization evaluation should move beyond asking whether outputs sound user-specific and instead ask whether LLMs make correct decisions for the users they represent. Code and resources are available at \url{https://github.com/xxy212/UserToolBench}.
\end{abstract}

\section{Introduction}

Tool-use LLMs are increasingly expected to act as personal assistants that search, plan, coordinate services, and invoke external tools on behalf of users.
In these settings, personalization is not only a matter of producing user-specific responses, but also of making correct personalized decisions for a persistent user.
A user request may omit decision-critical constraints such as budget, location preference, scheduling habits, or service style.
A reliable assistant must recover such preferences from prior interactions, decide whether clarification is needed, and produce a tool-call trajectory aligned with the user's established behavior.
We refer to this capability as \emph{personalized  decision making}.

Existing benchmarks provide important foundations for this problem.
Personalization benchmarks evaluate profile-conditioned generation, recommendation, and response adaptation \citep{salemi-etal-2024-lamp,jiang2025know,zhao2025personalens,gao2024prelude,li2024personalizedrlhf,prose2025}, while tool-use benchmarks evaluate tool selection, argument construction, and API-based task completion \citep{li-etal-2023-api,qin2023toolllm,tang2023toolalpaca,yu2026wildtoolbench}.
Interactive benchmarks further introduce realistic environments and long-horizon task execution \citep{yao2024taubench,astra2026}, and recent personalized tool-use benchmarks connect user profiles or histories with tool invocation \citep{petoolbench2025,ptbench2025,cheng-etal-2025-toolspectrum,hao-etal-2025-evaluating}.
As summarized in Table~\ref{tab:benchmark_comparison}, however, these lines of work have not yet jointly evaluated four requirements central to personalized delegation: executable tool calling, persistent user profiles, long-horizon reasoning, and realistic user-agent interaction.
This leaves open whether an LLM can use interaction history to make executable decisions that are correct for the particular user it represents.

\begin{table*}[t]
\centering
\small
\setlength{\tabcolsep}{5pt}
\begin{tabular}{lcccc}
\toprule
\textbf{Benchmark} & \textbf{Tool Use} & \textbf{Fixed Profile} & \textbf{Long Horizon} & \textbf{Realistic Interaction} \\
\midrule
LaMP~\citep{salemi-etal-2024-lamp} 
& -- & \checkmark & -- & -- \\

API-Bank~\citep{li-etal-2023-api} 
& \checkmark & -- & -- & -- \\

ToolBench (ToolLLM)~\citep{qin2023toolllm} 
& \checkmark & -- & -- & -- \\

$\tau$-bench~\citep{yao2024taubench} 
& \checkmark & -- & \checkmark & \checkmark \\

PEToolBench~\citep{petoolbench2025} 
& \checkmark & -- & -- & -- \\

PTBench~\citep{ptbench2025} 
& \checkmark & \checkmark & -- & -- \\

ToolSpectrum~\citep{cheng-etal-2025-toolspectrum} 
& \checkmark & \checkmark & -- & -- \\

WildToolBench~\citep{yu2026wildtoolbench}
& \checkmark & -- & -- & \checkmark \\

ETAPP~\citep{hao-etal-2025-evaluating} 
& \checkmark & \checkmark & -- & -- \\

\benchmark (ours)
& \checkmark & \checkmark & \checkmark & \checkmark \\
\bottomrule
\end{tabular}
\caption{Comparison between \benchmark\ and representative personalization or tool-use benchmarks.
\textbf{Tool Use} indicates whether the benchmark evaluates executable tool/API calls.
\textbf{Fixed Profile} indicates whether tasks are grounded in a persistent user profile.
\textbf{Long Horizon} indicates whether evaluation requires reasoning over extended multi-turn or cross-topic interaction trajectories.
\textbf{Realistic Interaction} indicates whether the interaction trajectory is designed to reflect realistic user-agent communication styles rather than isolated fully specified instructions.}
\label{tab:benchmark_comparison}
\end{table*}

We introduce \benchmark, a benchmark for evaluating personalized  decision making in tool-use LLMs.
The core design of \benchmark\ is a \emph{profile-hidden} evaluation protocol.
Reference trajectories are constructed and validated with access to persistent user profiles, so the target decisions are grounded in stable user preferences.
During evaluation, the tested LLM does not observe the explicit profile.
It receives only the accumulated interaction history, the current request, and the available tool schemas, and must infer the user-specific constraints needed for action.

This profile-hidden setting operationalizes personalized delegation as history-grounded executable decision making.
Because the explicit profile is hidden, models cannot rely on direct profile copying; they must recover relevant preferences from prior interactions.
Because evaluation is based on tool-call trajectories, personalization is measured through tool choices, argument values, clarification behavior, and multi-step action sequences rather than surface-level textual adaptation.
Because decisions are situated in long-horizon interactions, the benchmark tests whether models can maintain behaviorally consistent user alignment over time.
Thus, \benchmark\ integrates preference inference under incomplete information, tool-use planning, and long-horizon personalization in a single evaluation setting.

\noindent\textbf{Contributions.}
This work makes three contributions.
First, we formulate personalized  decision making as an incomplete-information tool-use problem grounded in persistent user behavior, and introduce \benchmark\ to evaluate this setting.
Second, we propose a profile-hidden evaluation protocol that tests whether LLMs can infer latent user preferences from interaction history and express them through executable tool-call trajectories.
Third, we evaluate strong tool-use LLMs and show that current models still struggle with reliable personalized delegation, especially in multi-tool coordination, missing-constraint inference, and long-horizon behavioral consistency.

\section{Related Work}

\paragraph{Personalized language modeling.}
Personalized language modeling studies how models adapt generated text to individual users. LaMP evaluates personalized generation tasks such as citation prediction, news headline generation, and review generation \citep{salemi-etal-2024-lamp}. Other work examines dynamic profiling, profile-conditioned response generation, and preference-aware alignment \citep{jiang2025know,zhao2025personalens,gao2024prelude,li2024personalizedrlhf,prose2025}. These studies demonstrate the value of user-specific context, but their evaluations remain largely response-level and do not directly test executable decisions made on behalf of users.

\paragraph{Tool-use LLMs.}
Tool-use benchmarks evaluate whether LLMs can select tools, construct valid arguments, and complete multi-step tasks. API-Bank, ToolLLM, ToolAlpaca, and WildToolBench study API and tool invocation at increasing scale and realism \citep{li-etal-2023-api,qin2023toolllm,tang2023toolalpaca,yu2026wildtoolbench}. Interactive benchmarks such as $\tau$-bench and ASTRA-bench further evaluate agents in simulated service or application environments \citep{yao2024taubench,astra2026}. Mem2ActBench evaluates whether agents can actively apply long-term memory to tool selection and parameter grounding across interrupted interactions \citep{shen-etal-2026-mem2actbench}. These settings establish important tool-use and memory capabilities, but generally do not jointly test whether a full tool trajectory is sensitive to the latent preferences of a persistent user whose structured profile is hidden.

\paragraph{Personalized tool use and user-centered evaluation.}
Recent work connects personalization with tool use in LLM-based assistants. PEToolBench, PTBench, ToolSpectrum, and ETAPP evaluate personalized tool invocation or proactive tool-augmented agents under user profiles, histories, or environments \citep{petoolbench2025,ptbench2025,cheng-etal-2025-toolspectrum,hao-etal-2025-evaluating}. Other benchmarks extend personalization to web navigation, mobile environments, shopping, travel planning, recommendation, and simulation \citep{cai2025personalwab,fingertip2026,persona2web2026,agenticshop2026,prefix2026,meagent2026,travelplannerplus2024,pts2025,valuepilot2025,recbenchplus2026,shopsimulator2026,propersim2026}. \benchmark\ differs by focusing on profile-hidden preference inference, clarification behavior, preference-sensitive tool-call trajectories, and long-horizon behavioral consistency within a single personalized  decision-making setting for tool-use LLMs.

As summarized in Table~\ref{tab:benchmark_comparison}, existing benchmarks usually cover only part of this setting. Some evaluate personalization without tool execution, some evaluate tool calling without persistent profiles, and others lack long-horizon or realistic user-agent interaction trajectories. \benchmark\ is designed to evaluate all four aspects together.

\section{UserToolBench}

\subsection{Task Formulation}

We formulate personalized  decision making as a sequential tool-use problem under incomplete information. In this setting, an LLM acts on behalf of a user according to stable preferences, rather than simply executing a fully specified instruction.

Let $p \in \pcal$ denote a persistent user profile, $h \in \hcal$ the historical interaction trajectory, $q$ the current request, and $T \in \tcal$ the available tool ecosystem.

We separate profile-conditioned reference construction from profile-hidden evaluation.
References are generated with access to $p$, while the evaluated LLM observes only $(h,q,T)$ and must infer user-specific constraints from history.

Formally, the reference decision trajectory is generated under
\begin{equation}
y^\star = f_{\mathrm{ref}}(p, h, q, T),
\end{equation}
where $p$ is visible to the reference generator and validators.
The evaluated LLM produces
\begin{equation}
\hat{y} = f_{\theta}(h, q, T),
\end{equation}
where $p$ is hidden.
This asymmetry is central to the benchmark: a model cannot copy an explicit profile, but must recover relevant preferences from earlier interactions and apply them to the current  decision.

The predicted trajectory $\hat{y}=\{a_1,\dots,a_N\}$ may include clarification queries, tool invocations with arguments, interpretation of tool feedback, and final user-facing responses.

When tool use is required, the decision process may involve a sequence of interactions with the external environment,
\begin{equation}
\tau = \{a^{T}_1, e_1, a^{T}_2, e_2, \dots, a^{T}_S, e_S\},
\end{equation}
where $a^{T}_s$ denotes the tool action issued by the LLM at step $s$, and $e_s$ denotes the corresponding environmental feedback.
During reference construction, the final response is generated with access to the user profile, interaction history, current request, and accumulated environmental observations.
During evaluation, the tested LLM does not observe the user profile and must rely on the interaction history, the current request, the available tools, and its inferred user-specific preferences.

Because $q$ may omit decision-critical constraints, the LLM must choose whether to ask for clarification, infer missing constraints from $h$, or proceed with the available information.
This differs from conventional tool-use benchmarks that evaluate explicit instruction execution, since success here depends jointly on tool orchestration, preference inference, and behavioral consistency.

\subsection{Data Source and Synthesis Pipeline}

\begin{figure*}[t]
\centering
\includegraphics[width=0.95\textwidth,height=0.45\textheight]{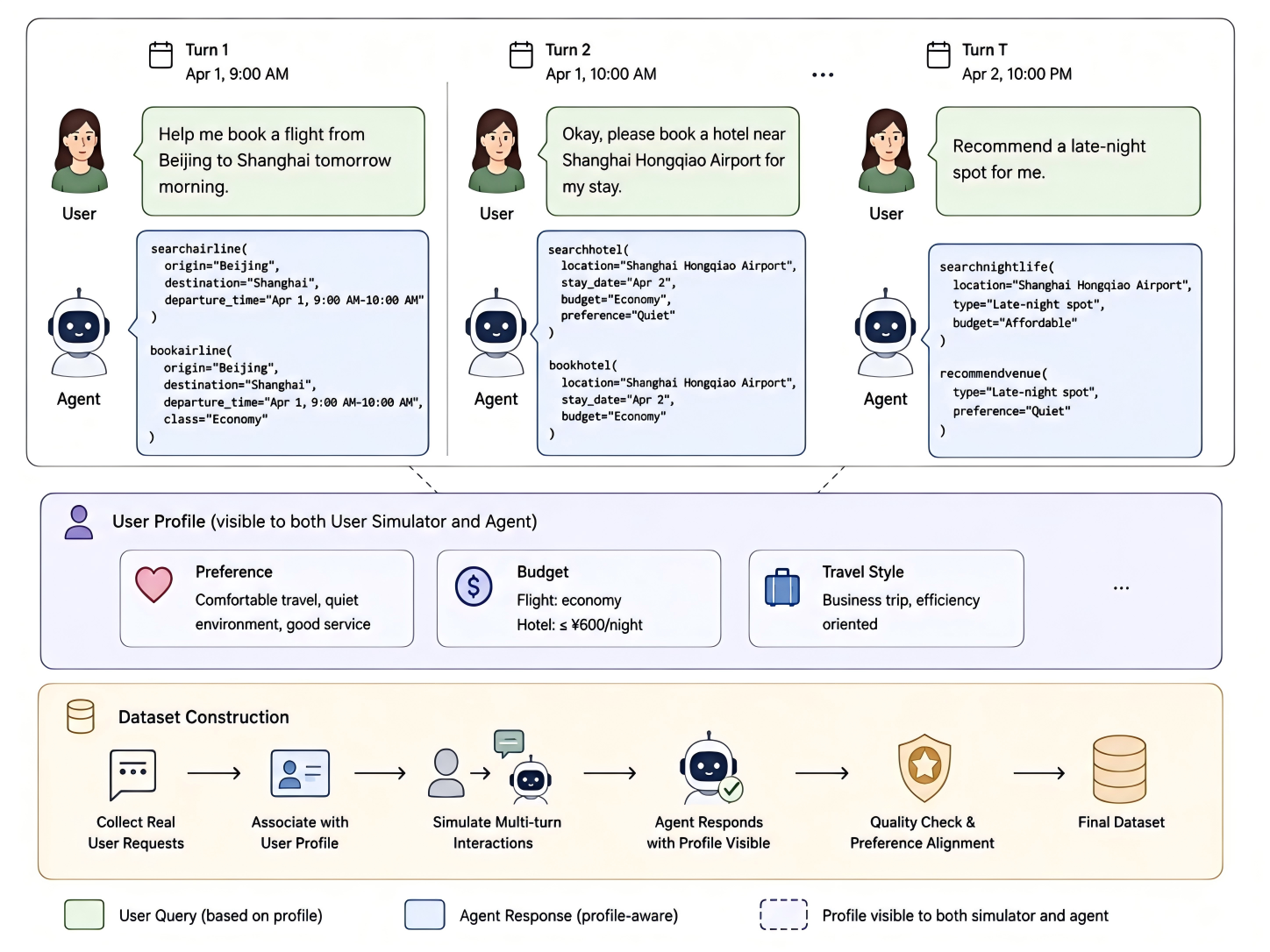}
\caption{
Overview of the \benchmark\ construction pipeline.
A persistent user profile is visible to both the user simulator and the reference trajectory generator during data synthesis.
The user simulator produces profile-consistent multi-turn requests, while the reference generator produces profile-aware tool-call trajectories.
}
\label{fig:pipeline}
\end{figure*}

We construct \benchmark\ with a profile-grounded data synthesis pipeline for realistic  decision making.
Each trajectory is tied to a persistent user identity: a persona-conditioned user simulator generates profile-consistent requests, and a reference trajectory generator produces profile-aware tool calls with access to the same profile.
Figure~\ref{fig:pipeline} illustrates the construction process and an example multi-turn trajectory.

\paragraph{User Profile Source.}
Each benchmark instance is associated with a persistent user profile built from privacy-sanitized real interaction traces.
Instead of retaining raw dialogues, we use an LLM-assisted abstraction process to convert traces into structured persona representations containing demographic descriptors, language background, communication style, personality traits, preference categories, and recurring task-relevant constraints.
Sensitive or identifying details are removed, generalized, or abstracted when unnecessary for task construction.
During generation and validation, we focus on non-identifying decision-relevant signals such as budget habits, travel style, scheduling conventions, service preferences, and communication norms.
Details of persona profile selection, schema, and privacy handling are provided in Appendix~\ref{app:persona-profile-construction}.

\paragraph{Tool Ecosystem Source.}
We construct tool environments from public API-style capabilities following prior tool-use benchmark construction practices such as ToolAlpaca~\citep{tang2023toolalpaca}.
Each tool is represented by a normalized function schema with a tool name, argument fields, type constraints, and functional description.
We manually organize tools into scenario-level ecosystems that match personal-assistant settings and contain meaningful decision points where user preferences can affect tool choices, tool order, or argument values.
Details are provided in Appendix~\ref{app:toolset-construction}.

\begin{table*}[t]
\centering
\small

\begin{subtable}[t]{0.47\textwidth}
\centering
\begin{tabular}{l r r}
\toprule
\textbf{Task Type} & \textbf{Count} & \textbf{Rate} \\
\midrule
Lack-of-Information & 278 & 34.79\% \\
Single-Tool & 265 & 33.17\% \\
Multi-Tool & 256 & 32.04\% \\
\bottomrule
\end{tabular}
\caption{Distribution of evaluation task types in \benchmark.}
\label{tab:task-types}
\end{subtable}
\hfill
\begin{subtable}[t]{0.47\textwidth}
\centering
\begin{tabular}{l r r}
\toprule
\textbf{Turn Subtype} & \textbf{Count} & \textbf{Rate} \\
\midrule
Cross-Topic & 300 & 28.17\% \\
Long-Range Dependency & 180 & 16.90\% \\
Partial Information & 585 & 54.93\% \\
\bottomrule
\end{tabular}
\caption{Distribution of turn subtypes in \benchmark.}
\label{tab:turn-subtypes}
\end{subtable}

\caption{Dataset statistics of \benchmark. Left: evaluation task types. Right: turn-level subtypes.}
\label{tab:dataset-statistics}
\end{table*}

\paragraph{Trajectory Collection.}

We collect interaction trajectories through a milestone-based synthesis procedure.
Given a persistent user profile and a scenario-level tool ecosystem, we first generate a set of profile-conditioned milestones.
Each milestone corresponds to a meaningful decision point within a topic, including the simulated user request, the structured reference tool-call trajectory, and the resulting tool observations.
The user simulator is conditioned on the profile and is instructed to issue natural requests that reflect the user's preferences, constraints, and communication style.
The reference trajectory generator is implemented as the planner component of the assistant pipeline. It is also given access to the same profile and the available tool schemas, and is required to produce executable tool-call plans rather than free-form conversational replies whenever external actions are needed.

This construction deliberately creates incomplete-information decision settings.
The simulated user is not required to restate all decision-critical constraints in every request.
Missing constraints may instead be recoverable from the user profile or from preceding milestones.
For example, if a user asks for a hotel near an airport, the request may omit the preferred budget or service style, and the reference trajectory must fill those arguments from the user's stable preferences.
For each topic, we use both LLM-based checking and human verification to ensure that the proposed tasks are consistent with the user profile, the selected tools and arguments are operationally valid, and the resulting decisions align with the user's preferences.
Compact role specifications for the user simulator, assistant LLM, and checker are provided in Appendix~\ref{app:multi-agent-prompts}; the full executable prompts are released at \url{https://github.com/xxy212/UserToolBench}.

To construct long-horizon cross-topic examples, we generate topic-level task scenarios in a milestone-by-milestone manner for each user.
Each milestone simulates a user request about one topic and the corresponding assistant interaction trajectory.
After every four milestones, the verified records from preceding milestones are incorporated into the interaction history for subsequent milestone generation.
This encourages the user simulator to create new requests that depend on earlier task contexts.
As a result, the benchmark includes not only within-topic personalization, but also cross-topic dependency and long-range preference consistency.

\paragraph{Final Dataset Format.}
Each example contains a multi-turn interaction context, the available tool ecosystem, and a reference decision trajectory represented mainly as structured tool calls. 
This format supports direct evaluation of both tool-use correctness and preference-aligned decision behavior.

\subsection{Dataset Statistics}

\benchmark\ contains 10 user profiles, covering 300 deduplicated topics and 1,065 turns in total.
Across these trajectories, the benchmark involves 170 unique tool names, indicating a diverse tool ecosystem rather than a narrow set of repeated API calls.

Table~\ref{tab:task-types} reports the distribution of 799 evaluation task instances by task type. These labels are assigned at the task-instance level rather than the dialogue-turn level; the full benchmark contains 1,065 dialogue turns.


We further annotate the full set of dialogue turns by turn-level subtype, as shown in Table~\ref{tab:turn-subtypes}.
Partial-information turns account for 54.93\% of the dataset, making incomplete user specification a central characteristic of the benchmark.
Cross-topic turns account for 28.17\%, requiring LLMs to connect the current request with information from other topics.
Long-range dependency turns account for 16.90\%, evaluating whether LLMs can use earlier interaction records when making later decisions.
Together, these annotations reflect the core goal of \benchmark\ : evaluating personalized  decision making under incomplete information and long-horizon interaction context.


\begin{figure*}[t]
\centering

\begin{subfigure}[t]{0.48\textwidth}
\centering
\includegraphics[width=\linewidth]{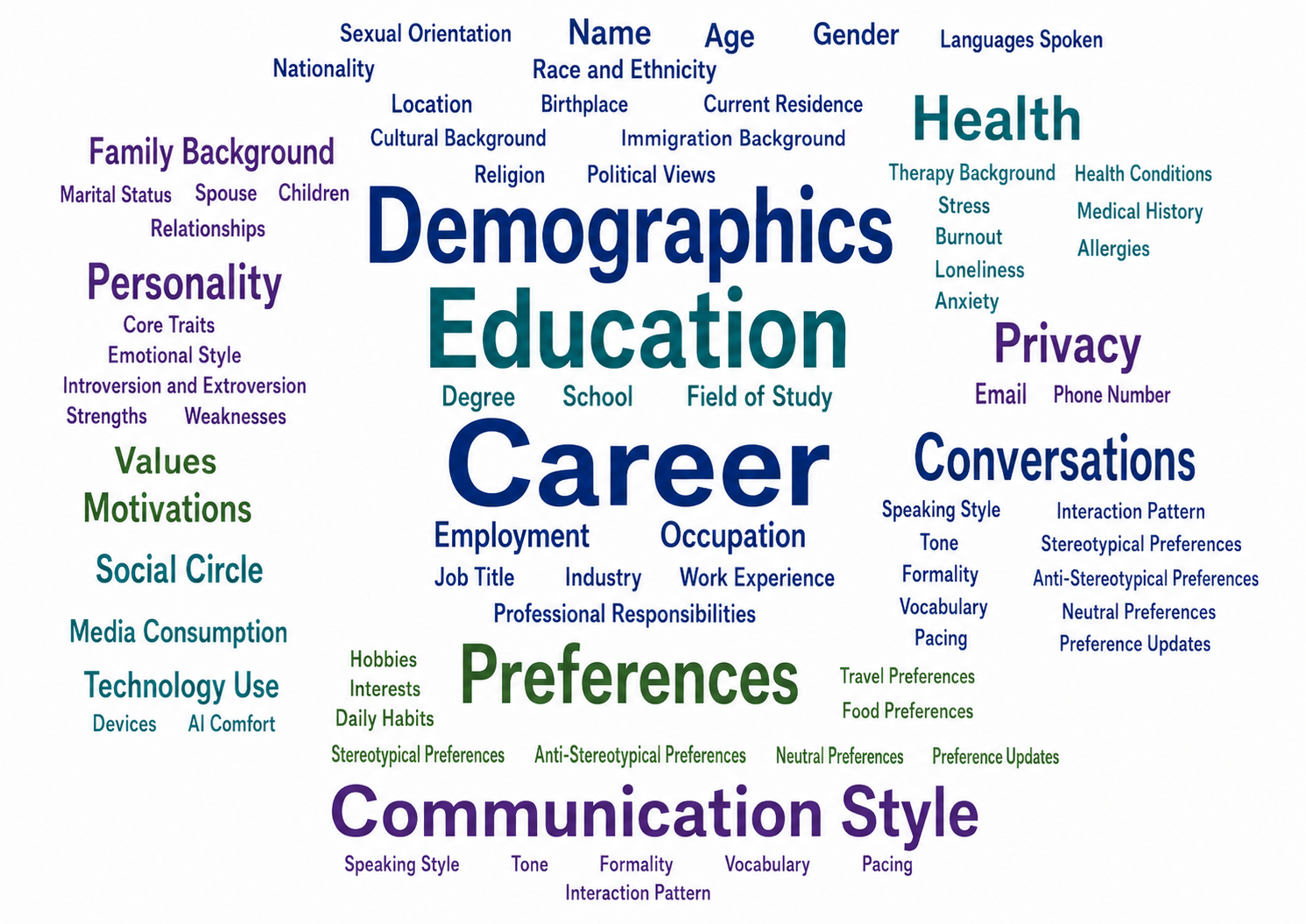}
\caption{
Persona profile field composition in \benchmark.
}
\label{fig:profile-field-composition}
\end{subfigure}
\hfill
\begin{subfigure}[t]{0.48\textwidth}
\centering
\includegraphics[width=\linewidth]{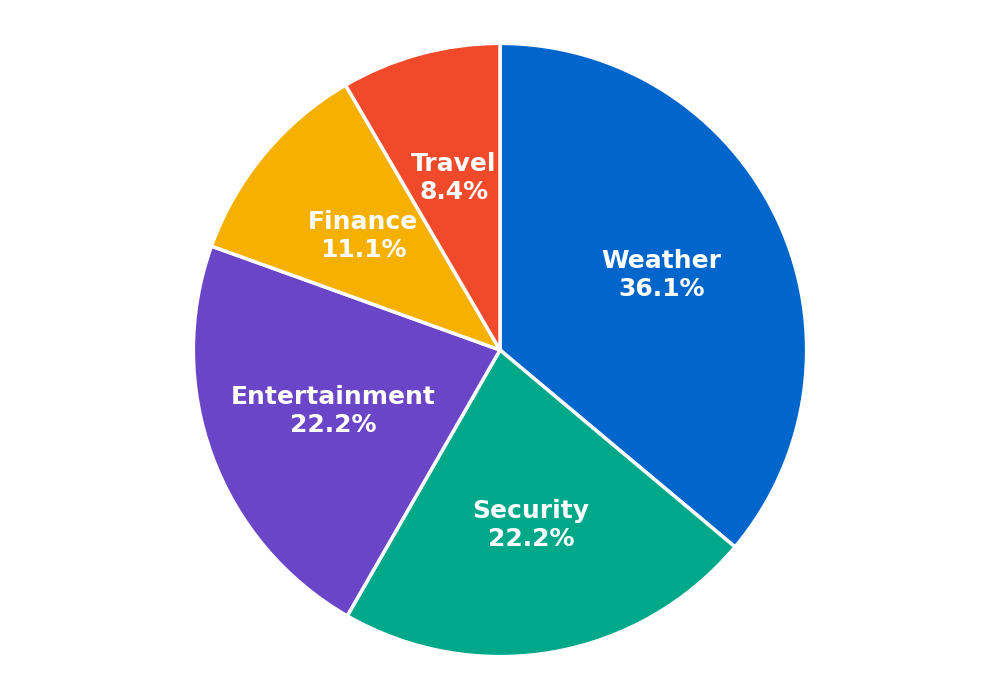}
\caption{
Scenario-level tool-domain distribution in \benchmark.
}
\label{fig:tool-domain-distribution}
\end{subfigure}

\caption{
Dataset composition of \benchmark.
Left: major field groups covered by structured persona profiles.
Right: main tool-domain categories used to construct personalized tool-use trajectories.
}
\label{fig:dataset-composition}
\end{figure*}

Figures~\ref{fig:profile-field-composition} and~\ref{fig:tool-domain-distribution} further visualize the composition of \benchmark.
Figure~\ref{fig:profile-field-composition} summarizes the major field groups appearing in the structured persona profiles, showing that the benchmark covers not only basic demographic attributes but also education, career, communication style, preferences, health-related context, privacy-related fields, and social or technology-use information.


Figure~\ref{fig:tool-domain-distribution} reports the distribution of scenario-level domains in the tool ecosystem, with weather, security, entertainment, finance, and travel forming the main categories.
Together, these visualizations show that \benchmark\ combines heterogeneous user-profile information with diverse assistant-oriented tool domains.


Following the general principle of testing whether persona conditions induce distinct observable behavior \citep{jiang-etal-2024-personallm}, we measure tool-call trajectory diversity across profiles under the same toolset.

\begin{table}[t] 
\centering 
\small 
\begin{tabular}{l r} 
\toprule \textbf{Statistic} & \textbf{Value} \\ \midrule 
Tool-sequence uniqueness ratio & 95.67\% \\ 
Exact-trajectory uniqueness ratio & 99.00\% \\ 
Avg. pairwise sequence distance & 72.77\% \\ 
Avg. pairwise exact-trajectory distance & 97.97\% \\
\bottomrule 
\end{tabular} 
\caption{Tool-call trajectory diversity across user profiles under the same toolset. } 
\label{tab:tool-call-diversity} 
\end{table}

\subsection{Evaluation Protocol}

\paragraph{Profile-hidden evaluation.}
A key feature of \benchmark\ is the asymmetry between data construction and model evaluation.
Reference trajectories are constructed with access to the persistent user profile and are manually verified for task validity and preference consistency.
In contrast, evaluated LLMs are not given the explicit profile.
They receive only the accumulated interaction history, the current user request, and the available tool schemas.
This setting tests whether an LLM can recover missing user-specific constraints from prior interactions and apply them when making  decisions through tool calls.

\paragraph{Tool-call correctness as  decision alignment.}
In \benchmark, user-specific preferences are grounded in executable tool decisions.
For underspecified requests, different users may require different tools, argument values, or clarification behavior even when the surface task is similar.

With reference trajectories constructed using validated user profiles and each processed toolset corresponding to a topic-level task setting, the same-toolset diversity reported in Table~\ref{tab:tool-call-diversity} reflects substantial profile-induced decision diversity across comparable topic instances. Therefore, exact tool-call correctness is a high-precision operational proxy for personalized decision alignment in our benchmark.
It should not be read as the only valid way to complete a user request.
Rather, it measures whether an evaluated LLM recovers one profile-conditioned reference decision path generated and verified using the persistent user profile.
Because most instances contain a single verified reference, exact matching can penalize an alternative trajectory that is also preference-compatible, and a binary mismatch alone does not indicate whether the deviation is minor or constitutes a substantial profile violation.
We therefore interpret Exact Acc. jointly with Relaxed Acc. and the multi-label and severity-aware diagnostics in Appendix~\ref{app:additional-diagnostics}. Developing multi-reference or equivalence-class evaluation for alternative user-aligned trajectories remains an important open problem.

\paragraph{Metrics.}
We report two complementary evaluation levels.

\paragraph{Exact trajectory accuracy.}
For each turn, we compare the predicted tool-call trajectory $\hat{y}$ with the reference trajectory $y^\star$.
For single-tool tasks, correctness requires matching the tool name and all required decision-relevant arguments.
For multi-tool tasks, correctness requires matching the ordered tool-call sequence and required arguments.
For lack-of-information tasks, correctness requires the LLM to ask for clarification when the missing constraint is unrecoverable from history, or infer it when recoverable from prior user behavior.
This strict metric assesses whether an LLM can reproduce the profile-conditioned reference decision trajectory.

\paragraph{Relaxed task completion accuracy.}
We also report Relaxed Task Completion Accuracy (Relaxed Acc.) to diagnose operational executability under relaxed trajectory matching.
A prediction is correct if it invokes valid tools from the available tool ecosystem, produces well-formed executable calls, and reaches a task-complete outcome under the explicit request and recoverable unambiguous constraints.
For multi-tool tasks, Relaxed Acc. credits alternative task-complete tool orderings.
For lack-of-information tasks, it credits appropriate clarification behavior or valid inference only when the missing constraint is unambiguously recoverable from the interaction history.
We report Relaxed Acc. using the same task-type and trajectory-position partitions as exact trajectory accuracy.

\section{Experiments}

\subsection{Experimental Setup}

We evaluate nine representative tool-use LLMs under the same profile-hidden setting.
The evaluated models include six strong general-purpose or frontier tool-use LLMs:
Kimi K2.6~\citep{moonshot2026kimi26},
GPT-5.4~\citep{openai2026gpt54},
Qwen 3.6 Plus~\citep{qwen2026qwen36plus},
DeepSeek V4 Pro~\citep{deepseekai2026deepseekv4},
Gemini 3.5 Flash~\citep{google2026gemini35flash},
and GLM-5~\citep{zai2026glm5}.
To provide additional comparison with smaller tool-specialized models, we also evaluate three tool-oriented models:
Hammer2.1-7B~\citep{lin2024hammer,madeagents2024hammer21},
ToolACE-2.5-8B~\citep{liu2025toolace,teamace2024toolace25},
and Watt-Tool-8B~\citep{wattai2024watttool8b,shi2024directmultiturnpreferenceoptimization}.

\subsection{Results}

Table~\ref{tab:overall-results} presents overall benchmark performance.
\begin{table*}[h]
\setlength{\tabcolsep}{4.5pt}
\centering
\begin{tabular}{lccccccc}
\toprule
Model & ST & MT & LOI & First-Third & Mid-Third & Last-Third & Avg. Exact Acc.  \\
\midrule

Kimi K2.6 & 52.30 & 26.44 & 37.20 & 44.24 & 43.41 & 33.67 & 38.65 \\
GPT-5.4 & 40.22 & 13.48 & \textbf{42.22} &  \textbf{52.12} & 40.93 & 31.33 & 31.98 \\
Qwen 3.6 Plus & 58.02 & 33.51 & 19.52 &  45.93 & 34.20 & 28.56 & 37.02 \\
DeepSeek V4 Pro & 61.95 & 33.26 & 32.29 & 43.59 & 44.26 & 39.15 & 42.49 \\
Gemini 3.5 Flash & 52.24 & 30.73 & 19.99 & 30.43 & 44.55 & \textbf{45.23} & 34.32 \\

GLM-5 & \textbf{70.62} & \textbf{38.41} & 39.07 & 49.35 & \textbf{52.64} & 43.90 & \textbf{49.36} \\
\midrule
Hammer2.1-7B & 38.50 & 17.16 & 18.61 & 27.05 & 25.16 & 22.75 & 24.76 \\
ToolACE-2.5-8B & 48.18 & 19.11 & 21.49 & 36.93 & 29.50 & 19.84 & 29.59 \\
Watt-Tool-8B & 39.66 & 15.08 & 7.66 & 26.59 & 21.31 & 13.59 & 20.80 \\
\bottomrule
\end{tabular}
\caption{
Exact trajectory accuracy on \benchmark\ by task type and trajectory position.
}
\label{tab:overall-results}
\end{table*}

\begin{table*}[t]
\setlength{\tabcolsep}{4pt}
\centering
\begin{tabular}{lccccccc}
\toprule
Model & ST & MT & LOI & First-Third & Mid-Third & Last-Third & Avg. Relaxed Acc. \\
\midrule
Kimi K2.6 & 78.00 & 61.29 & 58.09 & 72.45 & 72.50 & 51.23 & 65.79 \\
GPT-5.4 & 79.78 & 58.29 & \textbf{65.05} & \textbf{78.53} & 66.80 & 60.35 & 67.71 \\
Qwen 3.6 Plus & 82.39 & 71.26 & 40.79 & 68.65 & 69.21 & 50.63 & 64.81 \\
DeepSeek V4 Pro & \textbf{88.17} & 75.65 & 53.84 & 73.77 & \textbf{74.43} & \textbf{68.79} & \textbf{72.55} \\
Gemini 3.5 Flash & 79.42 & \textbf{76.77} & 35.88 & 62.36 & 71.68 & 67.30 & 64.02 \\
GLM-5 & 87.94 & 67.77 & 53.47 & 76.15 & 72.49 & 61.72 & 69.73 \\

\midrule
Hammer2.1-7B & 62.22 & 45.99 & 24.76 & 46.15 & 45.94 & 40.64 & 44.32 \\
ToolACE-2.5-8B & 70.27 & 45.85 & 30.71 & 62.26 & 43.56 & 36.91 & 48.94 \\
Watt-Tool-8B & 61.30 & 42.99 & 13.18 & 51.38 & 37.52 & 25.62 & 39.16 \\

\bottomrule
\end{tabular}
\caption{
Relaxed Task Completion Accuracy (Relaxed Acc.) by task type and trajectory position.
}
\label{tab:relaxed-task-completion}
\end{table*}

\paragraph{Current tool-use LLMs still struggle with strict personalized delegation.}
Table~\ref{tab:overall-results} shows that strict profile-conditioned trajectory matching is challenging across all evaluated models.
Even the best-performing model reaches only $49.36\%$ Avg. Exact Acc., showing that no tested model can reliably recover the profile-conditioned reference trajectory when the explicit user profile is hidden.
Although stronger general-purpose models generally outperform smaller tool-oriented models, the overall performance gap suggests that personalized delegation requires more than following the current instruction or generating syntactically valid tool calls.
Models must identify decision-relevant interaction history, distinguish stable user preferences from incidental context, and apply those preferences to tool choices and argument values.
The results reveal a gap between general tool-use competence and the ability to act as a reliable delegate for a persistent user.

\paragraph{Executable task completion does not imply personalized decision alignment.}
Table~\ref{tab:relaxed-task-completion} shows a large gap between relaxed task completion and exact trajectory matching.
For example, DeepSeek V4 Pro achieves $72.55\%$ Avg. Relaxed Acc. but only $42.49\%$ Avg. Exact Acc.; GPT-5.4 shows an even larger gap, from $31.98\%$ exact accuracy to $67.71\%$ relaxed accuracy.
This indicates that many predictions are executable and can satisfy the surface-level request, yet still fail to reproduce the profile-conditioned decision path.
Such errors often do not stem from malformed tool calls or lack of API knowledge.
Instead, they reflect cases where the model chooses a generally plausible tool sequence, tool order, or argument value that is nevertheless misaligned with the user's established preferences or expected evidence-gathering strategy.
These results support the central motivation of \benchmark: evaluating personalized assistants requires measuring whether the model makes the right personalized decision for the particular user, not merely whether it can complete the generic task.
Appendix~\ref{app:qualitative-case-study} illustrates this distinction with a case where a model issues a plausible metadata call but skips the history-grounded search step needed to recover the missing URL.

\paragraph{Strict mismatches are frequently substantive rather than cosmetic.}
To distinguish harmless planning variation from personalized decision failure, we conduct two additional diagnostics (Appendix~\ref{app:additional-diagnostics}).
A multi-label analysis of failed trajectories finds that sequence/dependency errors occur in $82.1$--$96.5\%$ of failures across the seven analyzed models, while wrong-tool decisions occur in $48.2$--$90.3\%$ and user-constraint violations in $40.9$--$53.5\%$.
A complementary severity audit reports only $0.3$--$2.1$ percentage points of slight preference deviation, compared with $28.8$--$65.5$ percentage points of major decision deviation under the diagnostic rubric.
These analyses do not eliminate the single-reference limitation, but they show that many exact-match errors involve missing calls, wrong tools, violated constraints, or broken dependencies rather than merely an interchangeable valid ordering.

\paragraph{Multi-tool delegation remains a major bottleneck.}
Across models, multi-tool tasks are consistently harder than single-tool tasks under exact trajectory matching.
The average accuracy drops from $51.30\%$ on single-tool tasks to $25.24\%$ on multi-tool tasks, showing that the difficulty extends beyond choosing one correct API.
Multi-tool delegation compounds several error sources: the model must select appropriate tools, decide their order, pass intermediate results across calls, and keep user-specific constraints active throughout the sequence.
Small deviations in early steps can propagate to later calls, causing the final trajectory to diverge from the profile-conditioned reference even when individual calls appear reasonable.
The low MT performance therefore reflects a sequential decision-control problem under personalization constraints, making multi-step tool orchestration a central bottleneck for reliable personalized delegation.

\paragraph{Preference inference under missing information is not captured by generic tool-use ability.}
LOI tasks reveal a capability distinct from ordinary tool execution.
In these cases, the model must first diagnose whether the user request is underspecified, then decide whether missing information can be safely inferred from prior behavior or should instead trigger clarification.
The results show that this ability does not necessarily track performance on explicit tool-use tasks.
For instance, GPT-5.4 obtains the highest LOI accuracy at $42.22\%$ despite its lower overall exact accuracy, whereas Qwen 3.6 Plus performs strongly on single-tool tasks at $58.02\%$ but drops to $19.52\%$ on LOI tasks.
This contrast suggests that missing-constraint handling requires uncertainty calibration as well as preference retrieval.
Over-inference may lead the model to act on unsupported assumptions, while excessive clarification may ignore preferences already recoverable from history.
Personalized delegation therefore depends on a separate ability to reason under incomplete information and judge when user history suffices for action.

\paragraph{Accumulated interaction history does not uniformly improve personalized delegation.}
The trajectory-position results show that longer histories do not automatically improve personalized decision making.
Although additional context can provide more evidence about user preferences, it also raises the burden of identifying relevant past interactions for the current request.
Several models degrade in later trajectory stages; for example, GPT-5.4 drops from $52.12\%$ in the first third to $31.33\%$ in the last third, and Qwen 3.6 Plus declines from $45.93\%$ to $28.56\%$.
This suggests that models may attend to stale, topic-specific, or incidental details while missing the stable preference signals that should guide the current decision.
However, the pattern is not universal: Gemini 3.5 Flash improves in later stages, suggesting that some models better exploit accumulated history.
The result highlights a key challenge for personalized assistants: history is useful only when the model can retrieve, filter, and update it selectively.
Long-horizon evaluation is therefore necessary because single-turn tasks cannot reveal whether models can maintain user alignment when useful signals are distributed across earlier interactions.

The main benchmark deliberately isolates stable preferences, but real assistants must also update user state when preferences change.
A preliminary dynamic-preference split constructed from 10 PersonaMem profiles encodes preference updates as temporally ordered interaction events and requires models to follow the most recent applicable preference using history alone (Appendix~\ref{app:dynamic-preferences}).
Average exact accuracy remains low---$30.08\%$ for GPT-5.4, $48.57\%$ for GLM-5, and $37.94\%$ for DeepSeek V4 Pro---indicating that preference updating is supported by the pipeline but remains challenging.

\section{Discussion}

\paragraph{Personalization over long interaction histories is a user-state tracking problem.}
The profile-hidden setting changes personalization from direct profile conditioning into selective inference over a growing behavioral record.
An agent must distinguish persistent preferences from incidental details, identify when a newer interaction supersedes an older preference, and retain uncertainty when evidence is incomplete or conflicting.
The stable-preference benchmark isolates the first of these capabilities, while the preliminary dynamic split demonstrates that the same construction framework can represent updates.
Systematic evaluation of preference reversals, conflicts among goals, temporary exceptions, and context-dependent priorities remains future work.

\paragraph{Exact matching is informative but not a complete utility function.}
A verified reference trajectory provides a precise and reproducible target for preference-sensitive tool choice, argument grounding, clarification, and dependency structure.
However, multiple trajectories can sometimes realize the same personalized intent, and binary matching does not express the severity of a deviation.
The relaxed metric, failure taxonomy, and severity-aware audit provide complementary views: they separate generic task completion from profile-conditioned alignment and distinguish mild argument-level variation from wrong tools, missing calls, violated constraints, and broken dependencies.
A stronger future protocol should represent sets of valid personalized trajectories or score semantic equivalence under explicit user constraints rather than relying on a single path.

\paragraph{Controllability and conversational realism remain in tension.}
LLM-assisted synthesis enables controlled coverage of profiles, tool ecosystems, missing-information cases, and long-range dependencies, while human verification filters invalid or unnatural instances.
Nevertheless, synthesized dialogue may retain prompt-specific lexical or discourse regularities and may not match authentic user--assistant conversations in repair behavior, hesitation, topic drift, or preference expression.
Consequently, benchmark performance should be interpreted as controlled evidence about personalized tool decisions, not as a complete estimate of deployment performance on unconstrained human dialogue.
Future validation should compare held-out authentic and synthesized interactions using blinded human judgments and distributional discourse measures.

\section{Conclusion}

We introduced \benchmark, a benchmark to evaluate personalized  decision making in tool-use LLMs. \benchmark tests whether models infer user preferences from interaction history and produce behaviorally aligned tool-call trajectories when explicit profiles are hidden and requests are incomplete. Experiments show current models still struggle with multi-tool coordination, missing-constraint inference, and long-horizon consistency. These results suggest personalization evaluation should move beyond whether responses sound user-specific and assess whether LLMs can make correct, executable personalized decisions for the users they represent.

\section{Limitations}
\benchmark\ has several limitations. First, the current version contains 10 user profiles, 300 deduplicated topics, and 1,065 turns. While this scale supports systematic analysis of profile-hidden personalized decision making, future work can further improve coverage by incorporating more user profiles, task domains, and interaction scenarios. Second, the benchmark uses simulated public API-style tool environments, which helps ensure controllability and reproducibility, but the tool ecosystem can be expanded to cover more real-world application settings. Finally, the current tasks mainly focus on personal-assistant-style tool-use scenarios. Future work can include more complex task compositions and richer tool dependencies to evaluate model behavior across a broader range of personalized decision-making settings.

\section{Ethical Considerations}

\benchmark\ is constructed from privacy-sanitized real interaction traces. We do not retain raw conversations in benchmark instances. Instead, interaction traces are abstracted into structured persona profiles, and sensitive or identifying details are removed, generalized, or replaced with non-identifying preference signals when they are not necessary for task construction. Personalized profiles may still encode sensitive behavioral patterns, so the benchmark should be released and used only in sanitized form.

\benchmark\ also highlights ethical risks in tool-use LLMs that make personalized decisions. LLMs that infer latent user preferences may make incorrect assumptions, over-infer sensitive attributes, reinforce stereotypes, or issue tool calls without sufficient user confirmation. Systems evaluated on \benchmark\ should therefore not be treated as ready for autonomous deployment in high-stakes domains such as healthcare, finance, legal services, or safety-critical decision making. Future work should report failure cases involving unsupported preference inference, inappropriate clarification behavior, and misuse of sensitive attributes.
\paragraph{Artifact license and intended use.}
We will release the sanitized benchmark artifacts of UserToolBench under the
Creative Commons Attribution-NonCommercial 4.0 International license
(CC BY-NC 4.0). The released artifacts are intended for non-commercial
research and educational use, including evaluation, analysis, and comparison
of tool-use LLMs under personalized  decision-making settings.
Raw interaction traces will not be redistributed. The released benchmark will
contain only sanitized and abstracted user profiles, tool schemas, interaction
contexts, and reference decision trajectories. Users of the benchmark should
not attempt to re-identify individuals, infer sensitive attributes beyond the
provided sanitized fields, or deploy systems trained or evaluated on this
benchmark for autonomous decision making in high-stakes domains such as
healthcare, finance, legal services, or safety-critical applications. If code
is released, it will be distributed separately under a permissive software
license such as the MIT License or Apache License 2.0.
\paragraph{Use of AI Assistants.}
We used large language models as research assistants during both manuscript preparation and dataset synthesis. In manuscript preparation, LLMs were used to support language polishing, clarity improvement, and organization of the writing. The authors reviewed, edited, and take full responsibility for all scientific claims, experimental results, analyses, and conclusions. In dataset construction, LLMs were used to assist with privacy-preserving profile abstraction, persona-conditioned user-request simulation, reference tool-call trajectory generation, and consistency checking. These model-generated components were subject to author inspection and validation to ensure that the resulting benchmark instances were consistent with the intended user profiles, tool schemas, and evaluation goals. No AI assistant was treated as an author, and the authors remain responsible for the integrity, correctness, and ethical handling of the submitted work.

\bibliographystyle{unsrtnat}
\bibliography{main}

\appendix
\section{Qualitative Case Study}
\label{app:qualitative-case-study}

This appendix provides a qualitative case study illustrating how \benchmark{} evaluates personalized  decision making beyond generic tool execution. The case focuses on a GPT model's behavior in a history-dependent media-intelligence task. Although the current user request is underspecified, the prior interaction history contains enough evidence to recover the intended article and construct a grounded tool-call trajectory.

\paragraph{Case context.}
The user is a founder/operator of Midwest Home Decor who uses the assistant for competitive retail intelligence, brand-narrative planning, and evidence-based media analysis. In the preceding interaction, the assistant discussed a Retail Daily article about GreenLeaf being named a retail rising star and noted that the story later received secondary coverage from 36Kr. The conversation then shifted to how the user could build a comparable non-sponsored authority signal for Midwest Home Decor, including a proposed ``Tuesday Retail Notes'' column centered on operational-efficiency topics such as sales per square foot.

The current user request is shown in Listing~\ref{lst:greenleaf-current-request}. The request asks for metadata of the previously mentioned 36Kr follow-up article, including the author, publication time, title rewrite angle, and editorial notes. Crucially, the user does not provide the article URL and instead expects the assistant to recover it from the prior context.

\begin{lstlisting}[
basicstyle=\ttfamily\scriptsize,
breaklines=true,
breakatwhitespace=false,
columns=fullflexible,
frame=single,
caption={Current user request. The request omits the URL and requires the assistant to recover the target article from history.},
label={lst:greenleaf-current-request}
]
I plan to make the first piece in the "Tuesday Retail Notes" series about sales per square foot,
but I need a benchmark example first. You just mentioned that the GreenLeaf article had
a secondary report on 36Kr. Could you pull the metadata for that news brief?
I want to examine its title rewrite angle and editorial notes. I did not save the URL
on my phone, but the system should be able to match it, right? Please first pull
the author and publication time for me.
\end{lstlisting}

\paragraph{History-dependent personalization signals.}
The correct decision is not determined by the current utterance alone. Several signals must be recovered from the interaction history. The request should be interpreted as business media intelligence rather than a generic news lookup, because the user is using the article as a benchmark for Midwest Home Decor's own brand-positioning strategy. The user also repeatedly distinguishes authoritative coverage from paid advertorials and asks for concrete metadata such as author, publication time, propagation path, rewrite angle, and editorial notes. This makes grounded evidence important: the assistant should not invent a URL merely because the requested source and topic appear obvious. The target entity and source are recoverable from the previous discussion: GreenLeaf is the competitor, and 36Kr is the secondary coverage source. The missing URL should therefore be obtained by searching for the history-derived query before calling the metadata tool.

\begin{table*}[t]
\centering
\small
\setlength{\tabcolsep}{4pt}
\begin{tabular}{p{0.20\textwidth}p{0.33\textwidth}p{0.39\textwidth}}
\toprule
\textbf{Signal} & \textbf{Evidence in history} & \textbf{Effect on the correct tool trajectory} \\
\midrule
User identity and goal & Founder/operator of Midwest Home Decor; wants Q3 strategy materials and authoritative brand coverage. & The assistant should treat the request as business media intelligence, not generic news lookup. \\
Preference for grounded evidence & The user repeatedly distinguishes authoritative coverage from paid advertorials and asks for author, publication time, propagation path, rewrite angle, and editorial notes. & The assistant should avoid filling missing URL arguments by guesswork; it must ground the URL through search. \\
Relevant prior entity & Previous turns identified GreenLeaf as the retail rising-star competitor and mentioned 36Kr secondary coverage. & The search query should combine the source and entity, e.g., ``36Kr GreenLeaf retail rising star''. \\
Immediate writing plan & The immediately preceding assistant proposed a ``Tuesday Retail Notes'' founder-column strategy, and the current user wants the first article to focus on sales per square foot. & The metadata is requested as a reusable writing template, so title rewrite and editorial notes are decision-relevant. \\
\bottomrule
\end{tabular}
\caption{Profile-relevant signals and interaction-history evidence for the selected case.}
\label{tab:greenleaf-personalization-signals}
\end{table*}

\paragraph{Relevant previous interaction.}
The current request depends on two preceding pieces of context. The earlier tool result established that the original Retail Daily report was followed by a 36Kr news brief, and the immediately preceding interaction introduced the user's ``Tuesday Retail Notes'' plan. Together, these turns explain why the user asks for title rewriting and editorial notes rather than only a URL.

\begin{lstlisting}[
basicstyle=\ttfamily\scriptsize,
breaklines=true,
breakatwhitespace=false,
columns=fullflexible,
frame=single,
caption={Relevant previous interaction snippets. These turns supply the missing article identity and the user's personalized writing objective.},
label={lst:greenleaf-previous-interaction}
]
Earlier assistant response:
Estimated propagation path and reach:
- Original platform: Retail Daily front page, approximately 45,000 impressions.
- Secondary repost: authorized repost by the WeChat account "Retail Business Review",
  approximately 28,000 impressions.
- Social amplification: reposted by key opinion leaders in LinkedIn industry groups,
  approximately 12,000 impressions.
- Media follow-up: secondary 36Kr news brief, approximately 65,000 impressions.

Immediately previous user turn:
Let's put aside the metadata of the GreenLeaf article for a moment.
I want to correct my request. If I want Midwest Home Decor to obtain
a comparable level of authoritative endorsement over the next two quarters,
instead of relying on paid advertorials for visibility, what three dimensions
should I use to reconstruct my brand narrative?

Immediately previous assistant response:
Dimension 3: contextualizing the founder's public voice from "business owner"
to "industry commentator."

Core action:
1. "Tuesday Retail Notes" LinkedIn column: publish every two weeks, with each
piece under 500 words, using anonymized operating data to comment on industry trends.
Example topics include "Why I no longer look at GMV and instead focus on sales
per square foot" and "How a 12-person team manages three stores and one online shop."
\end{lstlisting}

\paragraph{Reference trajectory.}
The reference trajectory first calls a search tool with history-derived keywords, then passes the returned URL to the metadata tool. This sequence converts an underspecified user request into a grounded executable trajectory.

\begin{lstlisting}[
basicstyle=\ttfamily\scriptsize,
breaklines=true,
breakatwhitespace=false,
columns=fullflexible,
frame=single,
caption={Reference tool-call trajectory. The assistant first grounds the missing URL through search, then retrieves article metadata.},
label={lst:greenleaf-reference-call}
]
searchArticles({
  "keywords": "36Kr GreenLeaf retail rising star",
  "limit": 10
})

Observation:
{
  "articles": [
    {
      "title": "GreenLeaf Fresh Named Retail Rising Star of the Year,
                Community Group-Buying Model Draws Industry Attention",
      "url": "https://36kr.com/p/2847291",
      "publishedAt": "2024-07-19T08:00:00Z",
      "source": "36Kr",
      "author": "Yang Lin"
    }
  ]
}

getArticleMetadata({
  "url": "https://36kr.com/p/2847291"
})

Observation:
{
  "title": "GreenLeaf Fresh Named Retail Rising Star of the Year,
            Community Group-Buying Model Draws Industry Attention",
  "author": {"name": "Yang Lin", "title": "36Kr consumer desk reporter"},
  "publish_date": "2024-07-19T08:00:00Z",
  "content_type": "news brief",
  "sponsored": false,
  "rewrite_angle": "Uses a financing-oriented angle, emphasizing growth data
                    and capital-market recognition.",
  "editorial_notes": {
    "data_source": "Retail Daily's original report and GreenLeaf's public financing information",
    "editor_remarks": "The title weakens rhetorical wording such as 'sweeping the market'
                       and instead uses 'draws industry attention' to maintain objectivity;
                       the brief adds financing information while preserving the core data points."
  }
}

prepare_to_answer({"answer_type": "tool"})
\end{lstlisting}

\paragraph{GPT prediction.}
The GPT model receives the long interaction history and the available tools, including both \texttt{searchArticles} and \texttt{getArticleMetadata}. However, instead of first grounding the missing URL through search, it directly calls the metadata tool with a plausible but unsupported URL.

\begin{lstlisting}[
basicstyle=\ttfamily\scriptsize,
breaklines=true,
breakatwhitespace=false,
columns=fullflexible,
frame=single,
caption={GPT prediction. The model skips the grounding step and fills the missing URL argument by guesswork.},
label={lst:greenleaf-gpt-prediction}
]
Available tools:
filterArticles, filterArticlesBySource, getArticleMetadata,
getHeadlines, getTrendingNews, searchArticles

GPT tool call:
getArticleMetadata({
  "url": "https://www.36kr.com/greenleaf-rising-star-2024"
})
\end{lstlisting}

\paragraph{Analysis.}
This case illustrates why executable-looking tool use is not sufficient for personalized  decision making. The predicted call is syntactically well formed and uses a relevant metadata tool, but it fails as a personalized decision because the missing URL is a decision-critical argument that should be grounded in the user's interaction history. The user's request combines a prior entity, a prior media source, and a personalized writing objective; therefore, the correct behavior is not to directly fill the URL slot, but to recover the article through \texttt{searchArticles} and then call \texttt{getArticleMetadata}.

The case also clarifies the distinction between surface-level task completion and profile-conditioned decision alignment. A generic assistant might treat the request as a simple metadata retrieval problem and attempt to guess the most likely article URL. In contrast, a reliable personalized delegate should preserve the user's preference for authoritative, non-sponsored evidence and avoid unsupported argument filling. Thus, the failure is not merely a tool-formatting error. It reflects a breakdown in history-dependent preference inference, uncertainty handling, and grounded tool-call planning, which are precisely the capabilities \benchmark{} is designed to evaluate.

\section{Persona Profile Construction and Field Coverage}
\label{app:persona-profile-construction}

\subsection{Candidate Profile Selection}

\benchmark\ uses structured persona profiles as persistent user states for constructing personalized  decision-making tasks.
We start from 100 candidate persona profiles derived from privacy-sanitized interaction evidence.
The current release selects 10 representative profiles through an LLM-assisted and human-verified filtering process.
The LLM is used to summarize candidate profile coverage and identify profiles with sufficiently rich decision-relevant preferences, while human annotators verify representativeness, internal consistency, privacy safety, and suitability for tool-based assistant scenarios.

The selected profiles share a common high-level schema, including persona summary, communication style, preference categories, task habits, and optional recurring constraints.
In addition to this shared schema, each profile may contain a small number of profile-specific fields.
These fields are retained only when they describe recurring, non-identifying, and task-relevant behavioral factors.

\subsection{Shared and Profile-Specific Schema}
\begin{table}[H]
\centering
\small
\begin{tabular}{p{0.28\linewidth} p{0.62\linewidth}}
\toprule
\textbf{Profile Field Group} & \textbf{Examples} \\
\midrule
Persona summary & short persona, personality, education, languages spoken \\
Communication & speaking style to chatbot, clarification preference \\
Preference categories & stereotypical, anti-stereotypical, and neutral preferences \\
Task habits & budget constraints, travel style, scheduling habits, service preferences \\
Optional fields & user-specific recurring constraints extracted from sanitized logs \\
\bottomrule
\end{tabular}
\caption{High-level schema of structured persona profiles. Sensitive or identifying details are abstracted or removed before benchmark construction.}
\label{tab:profile-schema}
\end{table}

Table~\ref{tab:profile-schema} summarizes the shared high-level schema used across persona profiles.
Table~\ref{tab:profile-specific-fields} reports the field coverage of the 10 selected profiles.
For privacy reasons, potentially sensitive profile-specific fields are reported at the category level rather than as raw field names.

\begin{table}[H]
\centering
\small
\begin{tabular}{lrrp{0.62\linewidth}}
\toprule
\textbf{Persona} & \textbf{Total Fields} & \textbf{Extra Fields} & \textbf{Profile-Specific Field Categories} \\
\midrule
Persona 1 & 22 & 2 & demographic background; occupation-related information \\
Persona 2 & 26 & 6 & geographic background; career information; family structure; residence; technology habits \\
Persona 3 & 29 & 9 & residence; daily-life narrative; family background; occupation; civic or belief-related context; socioeconomic context; technology usage; travel history \\
Persona 4 & 36 & 16 & geographic background; family structure; community roles; residence; former profession; hobbies and interests; economic context; migration history; occupation status; civic or belief-related context; technology comfort; travel history \\
Persona 5 & 29 & 9 & career information; family structure; location; hobbies and interests; economic context; technology preferences; travel history \\
Persona 6 & 30 & 10 & family structure; economic context; generational background; health-related context; hobbies and interests; location; occupation-related information; belief-related context \\
Persona 7 & 23 & 3 & career information; additional demographic context; location \\
Persona 8 & 29 & 9 & buying motivation; family structure; current goals; financial profile; location; marital or household context; occupation; psychographic profile; belief-related context \\
Persona 9 & 32 & 12 & geographic background; family structure; location; hobbies and interests; housing; economic context; occupation; demographic context; civic or belief-related context; household information \\
Persona 10 & 26 & 6 & location; demographic background; hobbies; musical background; technology use; work information \\
\bottomrule
\end{tabular}
\caption{Field coverage of the 10 selected persona profiles. All profiles share the high-level schema in Table~\ref{tab:profile-schema}. Profile-specific fields are summarized at the category level to avoid exposing sensitive or identifying details.}
\label{tab:profile-specific-fields}
\end{table}

\subsection{Quantitative Persona Diversity Audit}
\label{app:persona-diversity-audit}

We complement the field-coverage analysis with a quantitative audit of the ten selected profiles. Demographic descriptors span ages 22--60, with five female and five male profiles, seven nationality values, ten education backgrounds, and ten occupations. These counts characterize coverage in the selected sample; they do not imply representativeness of the broader user population.

\begin{table}[H]
\centering
\small
\begin{tabular}{lr}
\toprule
\textbf{Dimension} & \textbf{Coverage} \\
\midrule
Age & 22--60 years \\
Gender & 5 female, 5 male \\
Nationality & 7 unique values \\
Education & 10 unique values \\
Occupation & 10 unique values \\
\bottomrule
\end{tabular}
\caption{Summary coverage of selected persona descriptors.}
\label{tab:persona-demographic-coverage}
\end{table}

Following prior work on stable persona simulation, controllable persona generation, and lexical-diversity auditing \citep{luo-laban-2026-spasm,jiang-etal-2026-hachimi,kambhatla-etal-2025-measuring}, we quantify overlap both within matched profile fields and across complete non-sensitive persona text. For matched textual fields, we report the number of items, the fraction of unique entries, average TF--IDF cosine similarity, average token Jaccard similarity, corrected type--token ratio (CTTR), and moving-average type--token ratio (MATTR). Higher Unique, CTTR, and MATTR indicate less repetition or greater lexical diversity; lower TF--IDF and Jaccard values indicate lower cross-profile overlap.

\begin{table}[H]
\centering
\scriptsize
\setlength{\tabcolsep}{3.5pt}
\begin{tabular}{lrrrrrr}
\toprule
\textbf{Field} & \textbf{Items} & \textbf{Unique $\uparrow$} & \textbf{TF--IDF $\downarrow$} & \textbf{Jaccard $\downarrow$} & \textbf{CTTR $\uparrow$} & \textbf{MATTR $\uparrow$} \\
\midrule
Stereotypical preferences & 201 & 1.000 & 0.132 & 0.094 & 5.873 & 0.915 \\
Anti-stereotypical preferences & 275 & 0.993 & 0.176 & 0.103 & 7.223 & 0.946 \\
Neutral preferences & 284 & 0.958 & 0.299 & 0.176 & 6.463 & 0.904 \\
Therapy background & 200 & 1.000 & 0.150 & 0.121 & 8.465 & 0.960 \\
Speaking style to chatbot & 40 & 1.000 & 0.108 & 0.107 & 5.518 & 0.953 \\
\bottomrule
\end{tabular}
\caption{Field-level diversity of the selected persona profiles. Similarity values are lower-is-more-diverse; lexical-diversity values are higher-is-more-diverse.}
\label{tab:persona-field-diversity}
\end{table}

\begin{figure}[H]
\centering
\begin{minipage}{0.48\linewidth}
    \centering
    \includegraphics[width=\linewidth]{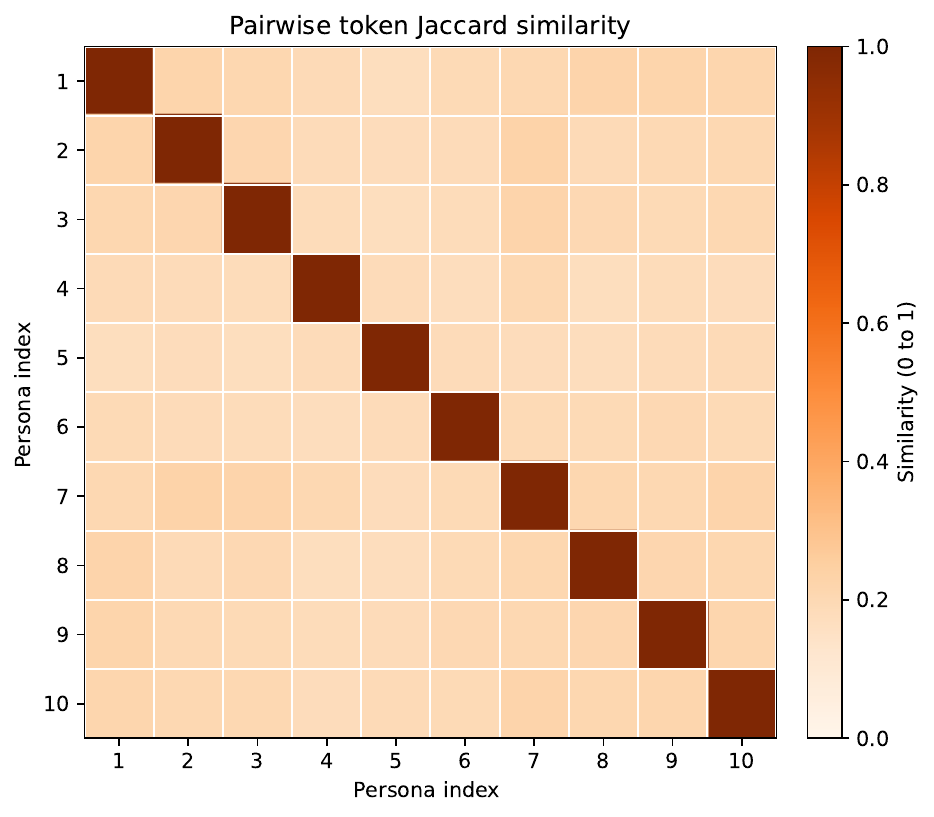}
\end{minipage}\hfill
\begin{minipage}{0.48\linewidth}
    \centering
    \includegraphics[width=\linewidth]{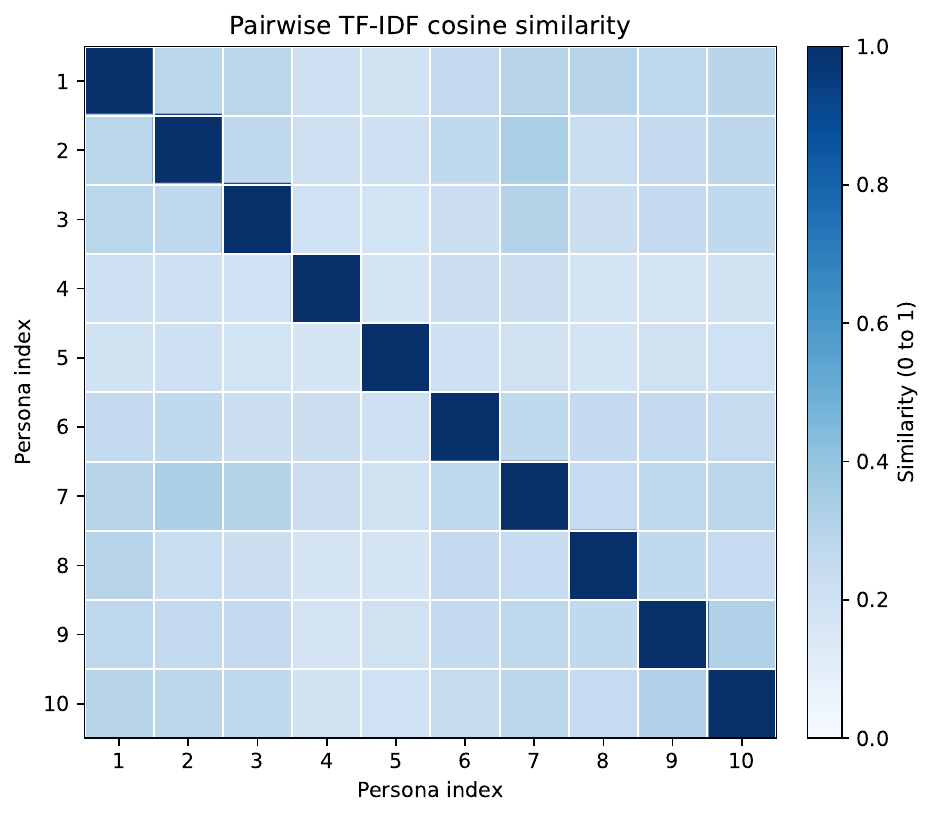}
\end{minipage}
\caption{Heatmap visualization of the complete pairwise full-profile similarity. Left: token Jaccard similarity. Right: TF--IDF cosine similarity. Lower values indicate greater diversity. Diagonal cells are masked because self-similarity is not part of the pairwise audit.}
\label{fig:persona-pairwise-diversity-heatmaps}
\end{figure}

At the full-profile level, we evaluate all $\binom{10}{2}=45$ persona pairs using only non-sensitive profile text. Across the 45 pairs, mean/median token Jaccard similarity is $0.197/0.194$ (range $0.169$--$0.228$), and mean/median TF--IDF cosine similarity is $0.242/0.250$ (range $0.172$--$0.338$). Persona 2 and Persona 7 form the most similar pair under both metrics, with Jaccard $0.228$ and TF--IDF cosine $0.338$.  Figure~\ref{fig:persona-pairwise-diversity-heatmaps} visualizes the complete pairwise audit as heatmaps rather than raw pair tables, fulfilling the complete pairwise audit while keeping the presentation compact.

Together with Table~\ref{tab:persona-field-diversity} and the trajectory-level diversity results in Table~\ref{tab:tool-call-diversity}, the pairwise heatmaps show that the selected profiles differ in both textual preference inventories and induced tool-call behavior. These results establish diversity within this controlled ten-profile sample, while not implying population-level representativeness.

\subsection{Privacy and Usage Constraints}

The selected persona profiles are behavioral abstractions rather than raw user records.
Raw conversations are not included in the released benchmark instances.
Sensitive or identifying details are removed, generalized, or replaced with non-identifying preference signals before task construction.
Profile fields are used to support  decision-making tasks only when they are relevant to user preferences, habits, communication style, or task constraints.

Potentially sensitive fields are not used to encourage demographic inference or stereotype-based personalization.
During task construction and validation, such fields are either generalized into non-identifying categories or excluded unless they are explicitly sanitized and necessary for a benign task constraint.
Tasks requiring unsupported inference over sensitive attributes are removed during validation.

\section{Tool Ecosystem Construction}
\label{app:toolset-construction}

\subsection{Raw Tool Collection and Scenario-Level Aggregation}

We construct the tool ecosystem in \benchmark\ from a large pool of API-style tool names.
The initial tool pool contains 1,248 tool-name occurrences.
Since many raw tools are overlapping, redundant, or too fine-grained to form coherent assistant tasks, we aggregate them into scenario-level toolsets rather than treating each tool independently.

Specifically, we first use an LLM-assisted clustering procedure to group semantically related tools into candidate toolsets.
Each toolset is intended to correspond to a realistic personal-assistant scenario, such as travel planning, weather searching, and so on.
The clustering stage produces 256 candidate toolset lines, covering 618 unique tool names.
On average, each candidate topic contains 4.875 tool names, and each tool contains 2.45 parameter slots.

\begin{table}[H]
\centering
\small
\begin{tabular}{lr}
\toprule
\textbf{Statistic} & \textbf{Value} \\
\midrule
Raw tool-name occurrences & 1,248 \\
Candidate toolset lines after LLM grouping & 256 \\
Unique tool names after grouping & 618 \\
Average tool names per topic & 4.875 \\
Average parameter slots per tool & 2.45 \\
Final manually selected toolsets & 36 \\
\bottomrule
\end{tabular}
\caption{Statistics of the tool ecosystem construction process. Raw API-style tools are first grouped into scenario-level candidate toolsets with LLM assistance and then manually filtered to retain coherent, preference-sensitive personal-assistant scenarios.}
\label{tab:toolset-construction}
\end{table}

\subsection{Manual Filtering}

The LLM-grouped candidate toolsets are manually filtered before inclusion in the benchmark.
The goal is not to maximize the number of tools, but to retain toolsets that support meaningful personalized decision making under user-specific preferences.
A retained toolset must form a coherent task scenario rather than a loose collection of unrelated APIs.
For example, a travel-planning toolset may include flight search, flight booking, hotel search, hotel booking, and local activity recommendation tools, because these tools naturally support a multi-step assistant workflow.

A retained toolset must also contain decision points that can be influenced by user preferences.
We remove toolsets whose outputs are almost entirely determined by explicit user instructions and leave little room for preference-sensitive decisions.
This criterion matters because \benchmark\ is designed to evaluate personalized  decision making rather than generic API invocation.

Finally, the toolset must be operationally usable in multi-turn trajectories.
We remove toolsets with underspecified functionality, incompatible argument schemas, excessive redundancy, or unclear execution dependencies.
When several candidate toolsets cover similar scenarios, we retain the one with clearer tool semantics and more diverse decision-relevant arguments.

After manual filtering, 36 distinct scenario-level toolsets remain in the final benchmark.
These toolsets constitute the available tool ecosystems used to construct and evaluate \benchmark\ trajectories.

\subsection{Decision-Relevant Tool Arguments}

For each retained tool, we additionally identify decision-relevant argument slots.
A decision-relevant slot is an argument whose value may change depending on the user's current request, prior interaction history, or stable preferences.
Examples include budget constraints, preferred location, time range, service style, ranking criterion, transportation mode, and accommodation preference.
In contrast, purely technical or formatting arguments, such as request identifiers, pagination size, or fixed output format, are not treated as personalization-sensitive decision slots.

This distinction is used during both data construction and evaluation.
During reference trajectory construction, preference-derived argument values are checked against the user profile and interaction history.
During evaluation, matching decision-relevant arguments is treated as stronger evidence of personalized  decision alignment than matching non-decision arguments.

\section{Additional Diagnostic Analyses}
\label{app:additional-diagnostics}

\subsection{Multi-Label Failure Analysis}
We align failed predicted trajectories with their profile-conditioned references and assign one or more diagnostic labels. \textit{Wrong Tool} denotes selection of a tool inconsistent with the intended personalized action; \textit{Missing Tool} denotes omission of a required call; \textit{User Constraint} denotes violation or loss of an explicit or history-grounded user constraint; \textit{Implicit Preference} denotes failure to recover a relevant latent preference; and \textit{Sequence/Dependency} denotes an incorrect ordering, missing dependency, or failure to pass information between calls. Because labels are multi-label, percentages within a row need not sum to $100\%$.

\begin{table}[H]
\centering
\small
\setlength{\tabcolsep}{5pt}
\begin{tabular}{lrrrrr}
\toprule
\textbf{Model} & \textbf{Wrong Tool} & \textbf{Missing Tool} & \textbf{User Constraint} & \textbf{Implicit Pref.} & \textbf{Seq./Dep.} \\
\midrule
Gemini 3.5 Flash & 79.5 & 16.5 & 41.7 & 11.0 & 84.3 \\
GLM-5 & 69.8 & 27.1 & 43.1 & 7.8 & 91.5 \\
GPT-5.4 & 48.2 & 45.6 & 53.5 & 6.1 & 96.5 \\
Hammer2.1-7B & 69.3 & 30.2 & 42.0 & 5.5 & 90.5 \\
Qwen 3.6 Plus & 73.3 & 21.7 & 45.0 & 5.0 & 91.7 \\
ToolACE-2.5-8B & 86.9 & 11.0 & 41.5 & 14.6 & 82.9 \\
Watt-Tool-8B & 90.3 & 8.2 & 40.9 & 16.6 & 82.1 \\
\bottomrule
\end{tabular}
\caption{Incidence (\%) of multi-label error types among failed trajectories for seven analyzed models.}
\label{tab:failure-analysis}
\end{table}

Sequence and dependency errors are the most frequent category for every analyzed model, indicating that personalized failures often propagate through multi-step execution rather than appearing as isolated formatting errors. Wrong-tool and user-constraint errors are also common, supporting the interpretation that many exact mismatches reflect substantive decision failures. Implicit-preference labels are less frequent because a latent preference error often becomes observable downstream as a wrong tool, omitted call, or violated constraint; the labels should therefore be interpreted jointly rather than as mutually exclusive causal categories.

\subsection{Severity-Aware Decision Diagnostics}
We further distinguish \textit{Slight Preference Deviation}, where the model selects the correct tool type but makes a limited departure from the optimal personalized decision, from \textit{Major Decision Deviation}, where it selects a substantially different tool path, omits a required call, or violates a user-specific constraint.

\begin{table}[H]
\centering
\small
\setlength{\tabcolsep}{6pt}
\begin{tabular}{lrr}
\toprule
\textbf{Model} & \textbf{Slight} & \textbf{Major} \\
\midrule
Gemini 3.5 Flash & 0.3 & 35.8 \\
GLM-5 & 0.8 & 32.9 \\
GPT-5.4 & 2.1 & 28.8 \\
Hammer2.1-7B & 0.4 & 62.5 \\
Qwen 3.6 Plus & 1.9 & 33.8 \\
ToolACE-2.5-8B & 1.4 & 55.9 \\
Watt-Tool-8B & 1.2 & 65.5 \\
\bottomrule
\end{tabular}
\caption{Reported diagnostic rates (\%) for slight and major personalized decision deviations.}
\label{tab:severity-diagnostic}
\end{table}

Across models, major deviations are substantially more frequent than slight deviations under this rubric. This result helps qualify the exact-match analysis: while some valid alternative trajectories may be penalized, the observed strict-performance gap cannot be attributed only to harmless preference-compatible variation. The diagnostic remains an auxiliary analysis rather than a replacement for multi-reference evaluation.

\subsection{Preliminary Dynamic-Preference Split}
\label{app:dynamic-preferences}
The main benchmark uses stable profiles to isolate history-grounded preference inference. To test whether the construction framework can represent evolving preferences, we additionally build a preliminary split from 10 PersonaMem profiles containing temporal preference updates \citep{jiang2025know}. Preference changes are represented as interaction events, and the evaluated model receives only the observed history, current request, and tools. It must follow the most recent applicable preference rather than the original profile state.

\begin{table}[H]
\centering
\small
\setlength{\tabcolsep}{4.5pt}
\begin{tabular}{lrrrrrrr}
\toprule
\textbf{Model} & \textbf{ST} & \textbf{MT} & \textbf{LOI} & \textbf{First-Third} & \textbf{Mid-Third} & \textbf{Last-Third} & \textbf{Avg. Exact} \\
\midrule
GPT-5.4 & 41.29 & 11.79 & 37.16 & 36.70 & 23.39 & 33.64 & 30.08 \\
GLM-5 & 63.14 & 26.07 & 56.50 & 61.93 & 35.64 & 53.51 & 48.57 \\
DeepSeek V4 Pro & 48.57 & 15.72 & 49.54 & 36.62 & 33.03 & 43.57 & 37.94 \\
\bottomrule
\end{tabular}
\caption{Exact trajectory accuracy (\%) on the preliminary dynamic-preference split.}
\label{tab:dynamic-preference-results}
\end{table}

The split confirms that the milestone-based pipeline can encode preference updates and evaluate whether a model follows the latest applicable state. Performance remains limited, particularly on multi-tool tasks, and all three models show a pronounced reduction in the middle third. These results should be interpreted as an extensibility study rather than a comprehensive benchmark of preference evolution; the split does not yet cover conflicts, reversals with uncertainty, or context-specific exceptions at scale.

\section{Compact Prompt Specifications for Trajectory Synthesis}
\label{app:multi-agent-prompts}

This appendix summarizes the role contracts used in the multi-role synthesis pipeline. We intentionally omit repeated formatting instructions and long enumerations that do not affect the methodological description. The complete executable prompts, including all construction modes and provider-specific wrappers, are released with the code at \url{https://github.com/xxy212/UserToolBench/tree/main/utb/generation/agent}.

The synthesis system uses a persona-conditioned user simulator, a planner, an assistant LLM, external tools, and checker roles. Instance-specific fields include the persona, dialogue history, environment information, tool schemas, required arguments, planner outputs, and tool observations. Explicit persona profiles are available only during data construction and validation; evaluated models receive interaction history rather than the structured profile.

\begin{table}[H]
\centering
\small
\setlength{\tabcolsep}{5pt}
\begin{tabular}{p{0.16\linewidth}p{0.34\linewidth}p{0.39\linewidth}}
\toprule
\textbf{Role} & \textbf{Primary inputs} & \textbf{Required behavior} \\
\midrule
User simulator & Persona, dialogue context, environment, tool schemas, construction mode & Generate a natural persona-consistent user turn; omit tool names; create complete, ambiguous, follow-up, inquiry-response, or non-tool turns as requested. \\
Planner & Current request, history, persona during synthesis, tool schemas & Decide whether to answer, clarify, or call tools; produce valid arguments and serial/parallel dependencies. \\
Assistant LLM & Planner output, tool observations, history, persona during synthesis & Ask only for unrecoverable information or produce a faithful user-facing response grounded in observations. \\
Checker & Request, history, environment, tool definitions, planner/tool outputs & Validate tool choice, required arguments, clarification need, execution dependencies, observation format, and persona consistency. \\
\bottomrule
\end{tabular}
\caption{Compact role specifications used for trajectory synthesis and validation.}
\label{tab:compact-role-prompts}
\end{table}

\begin{lstlisting}[
basicstyle=\ttfamily\scriptsize,
breaklines=true,
columns=fullflexible,
frame=single,
caption={Abbreviated role prompts. Full prompts are released in the repository.},
label={lst:compact-role-prompts}
]
USER SIMULATOR
Inputs: persona, history, environment, tools, generation mode.
Instruction: act as the same end user; express a natural task or reply
consistent with the persona. Do not name the intended tool. When the mode
is ambiguous, omit selected required information deliberately.
Output: User: <one dialogue turn>

ASSISTANT LLM
Inputs: planner decision, tool observations, history, persona, environment.
Instruction: follow the planner and observations faithfully. Ask a concise
question only when a required constraint cannot be recovered; otherwise
summarize results or answer directly in a persona-compatible style.
Output: Assistant: <user-facing reply>

CHECKER
Inputs: request, history, persona-conditioned context, environment, tools,
planner output, and tool observations.
Instruction: verify tool selection, required and decision-relevant arguments,
clarification behavior, serial/parallel dependencies, observation format,
and consistency with prior turns. Return a validity decision and reasons.
\end{lstlisting}

\subsection{Human Verification Interface}
\label{app:human_verification_interface}

To support human verification of the constructed trajectories, we built a lightweight web-based quality-control interface. The interface presents each benchmark instance together with its associated persona, dialogue messages, available tool set, and metadata. The goal of the interface is to help human verifiers inspect whether each trajectory is valid, profile-consistent, and safe to include in the benchmark.

The interface is organized into three panels. The left panel shows sample-level metadata, including the source file name, persona directory, toolset index, topic identifier, topic turns, node path, and environment information. It also displays a summarized persona. The full persona can be expanded when necessary, but the \texttt{sensitive\_information} field is hidden by default to reduce unnecessary exposure of sensitive content during verification.

The middle panel displays the complete message sequence for the current sample. Messages are visually separated by role, including user, assistant, tool, and other roles. This allows verifiers to inspect whether the conversation is coherent, whether the assistant follows the expected role protocol, and whether tool observations are used correctly in subsequent responses. If a sample cannot be parsed as valid JSON, the interface explicitly reports the parsing error.

The right panel displays the available tool set. For each tool, the interface shows the tool name, natural-language description, required arguments, and detailed parameter schema. Verifiers use this information to check whether the assistant selected the appropriate tool, supplied valid arguments, avoided hallucinated parameters, and asked for clarification when required information was missing.

For each sample, verifiers record one of three decisions: \textit{Pass}, \textit{Fail}, or \textit{Skip}. Failed samples can be assigned an issue type from a predefined taxonomy: incorrect tool selection, incorrect or hallucinated parameter, missing required clarification, multi-turn inconsistency, persona inconsistency, incorrect use of tool observations, role protocol error, sensitive information leakage, format or JSON error, or other. Verifiers may also provide free-form notes to explain the failure reason or document borderline cases.

The interface also provides a verification checklist. Verifiers are asked to examine whether the user request is natural and reasonably underspecified, whether the planner chooses the correct tool, whether missing required parameters trigger clarification, whether previously provided parameters are reused correctly, whether tool observations are consistent with the agent response, whether later turns properly follow from earlier turns, and whether the sample avoids unnecessary disclosure of sensitive information. This human verification step helps filter invalid, inconsistent, or privacy-risky examples before inclusion in the final benchmark.

\end{document}